\documentclass[letterpaper, 10pt, conference]{ieeeconf}
 \pdfoutput=1
\IEEEoverridecommandlockouts

\usepackage[utf8]{inputenc}
\usepackage[pdftex]{graphicx,color}
\usepackage{times}
\usepackage{amsmath}
\usepackage{amssymb}

\title{\LARGE
\textbf{Efficient Path Interpolation and Speed Profile
Computation for Nonholonomic Mobile Robots}}

\author{Stéphane Lens$^{1}$ and Bernard Boigelot$^{1}$%
\thanks{$^{1}$Montefiore Institute, B28, University of Liège, B-4000 Liège,
Belgium, \texttt{\small\{lens,boigelot\}@montefiore.ulg.ac.be}.}}

\newtheorem{theorem}{Theorem}

\begin{document}

\maketitle

\thispagestyle{empty}
\pagestyle{empty}

\begin{abstract}
This paper studies path synthesis for nonholonomic mobile robots moving in
two-dimensional space. We first address the problem of interpolating
paths expressed as sequences of straight line segments, such as those
produced by some planning algorithms, into smooth curves that can be
followed without stopping. Our solution has the advantage of being
simpler than other existing approaches, and has a low computational
cost that allows a real-time implementation.  It produces discretized
paths on which curvature and variation of curvature are bounded at all
points, and preserves obstacle clearance. Then, we consider the
problem of computing a time-optimal speed profile for such paths. We
introduce an algorithm that solves this problem in linear time, and
that is able to take into account a broader class of physical
constraints than other solutions. Our contributions have been implemented
and evaluated in the framework of the Eurobot contest.
\end{abstract}

\section{Introduction}

The general problem of moving a mobile robot as fast as possible from
a configuration to another while avoiding a given set of obstacles can
be tackled by several methods, such as cell
decomposition~\cite{brooks1985subdivision},
roadmap~\cite{karaman2011anytime}, and potential field
techniques~\cite{ge2002dynamic}.  In two-dimensional planar space, a
large number of planning algorithms produce paths that are expressed
as broken lines, i.e., sequences of straight line segments, between
the initial and final configurations.

A differential-drive robot cannot follow such a path without stopping
at the junction points between adjacent line segments in order to
change its orientation, which wastes time. This problem can be
alleviated by interpolating broken lines into smooth curves along
which the orientation of the robot and the curvature remain continuous
everywhere.  This paper first addresses the problem of computing such
interpolations, so as to obtain paths that can be physically followed
without the need for stopping or slowing down excessively. We consider
the case of differential-drive robots, but our results also apply to
tricycle or car-like platforms. We develop an interpolation algorithm
that guarantees that the obstacles cleared by a broken line are
avoided as well by the resulting smoothed out path.

Once an interpolated path has been obtained, we then address another
problem which consists in computing a time-optimal speed profile for
it, i.e., associating each point of the curve with a timestamp that
provides the instant at which it will be visited, in such a way that
the total time needed for following the path becomes as small as
possible.  This computation has to take into account various physical
constraints of the robot, e.g., bounds on its velocity or acceleration
measured at its wheels or its center of mass, limits imposed by the
steering mechanism, \ldots\ Some of these constraints may be
context-sensitive, e.g., a tight speed bound may be imposed in the
vicinity of obstacles cleared by a small margin, or steering
constraints may be expressed as a function of the robot velocity.

Our motivation for studying these problems originates from our
participation to the
Eurobot\footnote[2]{\texttt{http://www.eurobot.org}} contest, in which
autonomous mobile robots compete in a short-duration game played on a
$2 \times 3\,\mathrm{m}^2$ area. For this application, it is essential
to plan paths in real-time due to the dynamic nature of obstacles,
which practically requires a method with a computational cost limited
to milliseconds of CPU time, as well as to obtain trajectories that minimize
the time needed for moving the robot from one configuration to
another. This prompted the development of a time-optimal speed profile
computation algorithm that takes into account all the relevant
physical constraints, such as those limiting traction at the robot
wheels, or needed for ensuring stability in turns. Another requirement
that is specific to the Eurobot application is to receive detailed and
precise information about the locations that will be visited by the
robot and their associated timestamps before starting to follow a
trajectory, in order to be able to coordinate complex actions such as
actuations carried out when the robot is moving, or jointly performed
by two partner robots. While other approaches such
as~\cite{kunz12} and~\cite{kuwata09} have been proposed to
deal with some of these requirements, our solutions are, to the best
of our knowledge, the first ones that meet all of them in a
satisfactory way. Our algorithms have been implemented and evaluated
in the robots that we have built for Eurobot since 2008, which amounts
to hundreds of thousands of trajectories successfully synthesized.

\section{Problems Statement}

\subsection{Interpolation Problem}

The first problem that we consider consists of smoothing out a path
expressed as a broken line. We define a
\emph{broken line path} as a sequence $p_0, p_1, \ldots p_n$ of
points, with $n \geq 1$, such that
\begin{itemize}
\item
each point $p_i$, with $i \in [0, n]$ is defined by its coordinates
$(x_i, y_i)$ in two-dimensional space, and
\item
each intermediate point $p_i$, with $i \in [1, n-1]$, is associated
with a \emph{clearance parameter} $c_i \in \mathbb{R}_{> 0} \cup
\{ +\infty \}$.
\end{itemize}

\begin{figure}
\vspace{5pt}
\centerline{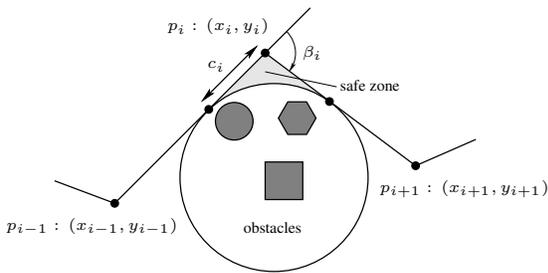}
\caption{Safe zone between adjacent segments}
\label{fig-segments-clearance}
\vspace{-5pt}
\end{figure}

The path is composed of the successive straight line segments
$[p_0, p_1]$, $[p_1, p_2]$, \ldots, $[p_{n-1}, p_n]$. In order to deal
with obstacles, we assume that robots have zero measure, in other
words, that a path clears a set of obstacles iff the
intersection between its line segments and the union of all these
obstacles is empty. (The case of robots with a cylindrical geometry
can straightforwardly be handled by dilating obstacles.) The purpose
of the clearance parameter $c_i$ is to provide additional information
about the location of the obstacles that are avoided when moving along
the segments $[p_{i-1}, p_i]$ and $[p_i, p_{i+1}]$. This is achieved
by considering a disk $D_i$ that is tangent to both segments (which
implies that its center belongs to the inner bisector of the angle
formed by these segments), and that fully covers the obstacles cleared
by the pair of segments. This latter property precisely means that the area
located between the disk $D_i$ and the segments $[p_{i-1}, p_i]$ and
$[p_i, p_{i+1}]$ is free from obstacles; we call this area the
\emph{safe zone} of this pair of segments. (By extension, we consider
that the segments themselves also belong to their safe zone.) 
The point is that any interpolation of the path that is confined to
safe zones is guaranteed to avoid obstacles. The safe zone across
$p_i$ is characterized by the parameter $c_i$,
defined as the distance between $p_i$ and the points of tangency
between $D_i$ and the segments $[p_{i-1}, p_i]$ and $[p_i,
  p_{i+1}]$. This parameter may be left undefined ($c_i = +\infty$),
in which case its value can be replaced by the smallest of the lengths
of the two segments. (In other words, $D_i$ is then the largest disk
simultaneously tangent to both segments.) An illustration is provided in
Figure~\ref{fig-segments-clearance}.

For each $i \in [1, n-1]$, we define $\beta_i$ as the angle between
the vectors $\overrightarrow{p_{i-1} p_i}$ and $\overrightarrow{p_i
  p_{i+1}}$, which corresponds to the change in orientation of the
robot when it moves from the segment $[p_{i-1}, p_i]$ to the segment
$[p_i, p_{i+1}]$. Without loss of generality, we assume $\beta_i \neq
0$. It is also natural to impose an upper bound on
$|\beta_i|$. Indeed, with a large value of $\beta_i$, the segments
$[p_{i-1}, p_i]$ and $[p_i, p_{i+1}]$ can be followed in opposite
directions, and it is not always appropriate in such cases to
interpolate the path into one that remains close to the segments. In
this work, we arbitrarily impose the upper bound $|\beta_i| \leq
\pi/2$ for all $i \in [1, n-1]$. In the case of adjacent segments
forming an acute angle, it then becomes necessary to interleave an
intermediate segment between them.

A differential-drive robot subject to physical constraints cannot
follow a broken line path without stopping at junction points. Such
constraints usually take the form of lower and upper bounds on
the velocity and acceleration of the robot measured at specific
locations, such as its individual wheels, center of mass, or
other reference centers. The speed that can be reached by the robot at
some point of a path is then bounded by a function of the
absolute curvature $|\kappa|$ at this point, as well as the rate of
variation $|d\kappa/ds|$ of this curvature with respect to the linear
traveled distance.

We are now ready to define precisely the interpolation problem. Given
a broken line path, the goal is to compute a curve that leads from its
origin to its endpoint staying within safe zones, and such that the
absolute curvature and variation of curvature remain small at all
points. For our intended applications, it is essential to be able to
carry out this operation with a low computational cost.  Finally, one
must be able to discretize the resulting interpolated path. This 
discretization must be physically sound, in the sense that the discretized 
values of the physical variables of interest (such as speeds and 
accelerations) must remain close to their actual value.

\subsection{Speed Profile Problem}
\label{sec-speed-profile-statement}

The second problem that we address takes as input a discretized path,
expressed as a sequence of configurations $(x_i, y_i, \theta_i)$
successively visited by a robot, where $(x_i, y_i)$ denotes the
two-dimensional coordinates of its reference center, and $\theta_i$
its absolute orientation. In addition, some number of physical
constraints that need to be satisfied at all times are provided, such
as lower and upper bounds on the velocity or acceleration of
individual wheels, the speed, angular speed, and tangential and radial
accelerations measured at the center of mass or some other reference
points, on the rate of variation of the steering angle for a tricycle
robot, \ldots\ Some of those constraints may be context-sensitive,
such as imposing tighter speed bounds in the vicinity of some
obstacles, or expressing the admissible angular velocity of the steering
wheel as a function of the robot speed. Besides those constraints, the
initial speed of the robot is specified at the origin of the path,
together with an upper bound on the speed that can be reached at its
endpoint. Given a path and a set of physical constraints, the aim is
to compute for all visited configurations $(x_i, y_i, \theta_i)$ a
timestamp $t_i$ that defines the time at which this configuration will
be reached, starting from $t_0 = 0$. The goal is to obtain the speed
profile that minimizes the total time needed for following the path,
while satisfying the physical constraints at all times.

\section{Path Interpolation}
\label{sec-interpolation}

We solve the interpolation problem in two steps, the first one being
aimed at producing a path in which the absolute curvature is bounded
at all points, and the second one modifying this path in order to
now bound the rate of variation of curvature. In both steps, the
interpolated path has to stay within safe zones in order to clear
obstacles.

\subsection{Bounding Curvature}

In order to bound absolute curvature, we build a curve composed of
straight line segments (with zero curvature) and circle arcs (with
constant curvature), connected in such a way that continuity of the
tangent vector is ensured everywhere.  On such a curve, the curvature
can be expressed as a piecewise constant function with respect to
traveled distance.

We construct such a curve by computing, for each pair of adjacent
segments $([p_{i-1}, p_i], [p_i, p_{i+1}])$ a value
$\ell_i$ specifying the distance from $p_i$ at which the curve
transitions from the segments to a circle arc. In other words, $\ell_i$
corresponds to the distance between $p_i$ and each point of tangency between
that circle arc and the segments.

Of course, in order to clear obstacles, it is necessary to have
$\ell_i \leq c_i$ for all $i \in [1, n-1]$. We compute $\ell_i$ by
applying the following principle: If three or more consecutive
segments are all tangent to a common circle, then the arcs that
interpolate these segments must belong to that circle, provided that
they are located within safe zones. This solution has the desirable
property of keeping the curvature constant across two or more
interpolation steps.

\begin{figure}
\vspace{5pt}
\centerline{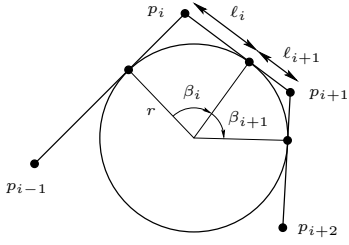}
\caption{Segments tangent to a common circle}
\label{fig-tangents}
\vspace{-5pt}
\end{figure}

We now show how to carry out this computation. Consider a path
in which all segments are tangent to a common circle of radius $r$.
This situation is illustrated in Figure~\ref{fig-tangents}.
At the points $p_i$ and $p_{i+1}$, one has respectively $\ell_i = r |\tan
(\beta_i / 2)|$ and $\ell_{i+1} = r |\tan (\beta_{i+1} / 2)|$. Since,
in this case, the constraint $\ell_i + \ell_{i+1} = |p_i p_{i+1}|$
is satisfied, we obtain
\begin{eqnarray}
\displaystyle
\ell_i & =
\displaystyle
\frac{\tau_i\,|p_i p_{i+1}|}
{\tau_i + \tau_{i+1}}\\
\displaystyle
\ell_{i+1} & =
\displaystyle
\frac{\tau_{i+1}\,|p_i p_{i+1}|}
{\tau_i + \tau_{i+1}},
\end{eqnarray}
where $\tau_i = |\tan(\beta_i/2)|$ for all $i$. Note that these expressions do
not involve $r$, and that Equation~(1) can be rewritten at the point
$p_i$ into
\begin{equation}
\ell_{i} =  \frac{\tau_{i}\,|p_{i-1} p_{i}|}
{\tau_{i-1} + \tau_{i}}.
\end{equation}

For general paths, successive segments are not tangent to
a common circle, and Equations~(1) and~(3) then provide different values for
$\ell_i$. Our strategy is, for all $i \in [1, n-1]$, to define $\ell_i$
as the smallest value among those expressed by Equations~(1) and~(3),
and the clearance parameter $c_i$. This solution also applies to
pairs of adjacent segments that turn in opposite directions; in such a case,
small values of $|\beta_i|$ (which represent small changes of direction)
lead to small circle arcs, and large values of $|\beta_i|$ to large arcs,
which is geometrically sound.

\subsection{Bounding Curvature Variations}

We now turn to the problem of modifying the path produced at the
previous step, which has a curvature that is piecewise constant, into
one in which the rate of variation of curvature with respect to
traveled distance remains bounded.  Clearing obstacles is achieved by
constraining the interpolation to remain within safe zones.  The
resulting path must have a curvature that is continuous, bounded, and
of bounded slope, at all points.

We construct such curves out of \emph{clothoids}, which correspond to
the paths followed by differential-drive robots when their wheels are
driven at respectively their minimum and maximum acceleration (as a
result, for instance, of bang-bang control). Clothoids are formally
defined as curves with a curvature that varies linearly with traveled
distance. The coordinates of the points visited by a clothoid are
expressed in terms of Fresnel integrals, which cannot be evaluated
analytically, but for which very efficient numerical approximations
are known~\cite{mielenz2000computation}.

Consider a circle arc with curvature $\kappa_C$, interpolating two
successive line segments $[p_{i-1}, p_i]$ and $[p_i, p_{i+1}]$ within
their safe zone. Assuming w.l.o.g.  $\kappa_C > 0$ (the case $\kappa_C
< 0$ is handled symmetrically), we have established the following
result.

\begin{theorem}
\label{theo-two-clothoids}
For every curvatures $\kappa_1, \kappa_2$ such that $0 \leq \kappa_1 <
\kappa_C$ and $0 \leq \kappa_2 < \kappa_C$, there exist two clothoids
arcs moving respectively from the curvatures $\kappa_1$ to $\kappa_M$
and from $\kappa_M$ to $\kappa_2$, with $\kappa_M > \kappa_C$, the
concatenation of which interpolates the path from $p_{i-1}$ to
$p_{i+1}$ within the safe zone, with continuity of the tangent
vector at the junction point between the two curves. The parameters
of these two clothoids arcs are uniquely determined by $\kappa_1$,
$\kappa_2$, $\kappa_C$, and the rotation angle $\beta_i$.
\end{theorem}

\begin{figure}
\vspace{5pt}
\centerline{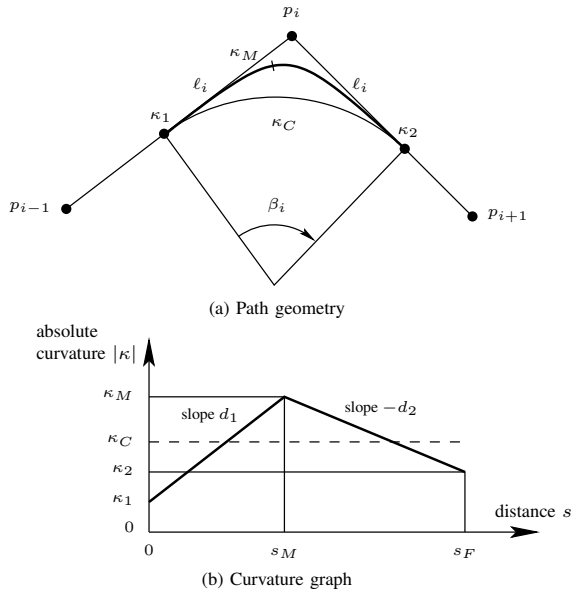}
\caption{Interpolation with two clothoid arcs}
\label{fig-clothoids}
\vspace{-5pt}
\end{figure}

Our method for characterizing the two clothoid arcs consists in
reasoning on a diagram expressing the curvature of the interpolated
path as a function of traveled distance. The problem is illustrated in
Figure~\ref{fig-clothoids} (exaggerating the curvatures in
order to make the interpolated path stand out from the circle
arc). Let $s_M$ denote the distance traveled along the first arc, and
$s_F$ the total distance traveled over both arcs. The linear rate of
variation of curvature for these arcs are respectively denoted by
$d_1$ and $-d_2$.  In the graph depicted in
Figure~\ref{fig-clothoids}(b), the area below the curvature line must
be equal to $\beta_i$.  Note that $s_M$ and $\kappa_M$ can be
expressed in terms of the other variables, since one has
$\kappa_M = \kappa_1 + d_1 s_M = \kappa_2 +
d_2 (s_F - s_M)$.

It thus remains to compute $d_1$, $d_2$
and $s_F$ given $\kappa_1$, $\kappa_2$ and $\beta_i$.
We solve this problem numerically, observing empirically that the
initial estimate
\[
\frac{s_F}{s_M} = 1 + \frac{\kappa_C - \kappa_1}{\kappa_C -
\kappa_2}
\]
leads to very fast convergence with Newton-Raphson's
method. In practice, we first perform a variable change operation by
defining
\[
a = \frac{d_1 - d_2}{d_1 + d_2}
\mbox{~~and~~}
d = \frac{d_1 + d_2}{2},
\]
and
then carry out the search over the variables $s_F$, $a$, and
$d$. Intuitively, $a$ is a measure of the asymmetry between the two
clothoids arcs, and remains small when $\kappa_1$ and $\kappa_2$ are
reasonably balanced. The value of $s_F$ is confined to an interval
with a lower bound ${\beta_i}/{\kappa_C}$
equal to the distance traveled
on the circle arc. Its upper bound corresponds to the minimum value
among
\[
\frac{2}{\kappa_C} \tan \frac{\beta_i}{2} \mbox{~~and~~}
\frac{\beta_i}{\min(\kappa_1, \kappa_2)},
\]
which respectively correspond to the combined length
of the line segments, and
the largest distance that can be traveled without lowering the
curvature below $\kappa_1$ and $\kappa_2$. It is also worth mentioning
that, in the particular case $\kappa_1 = \kappa_2 = 0$, this procedure can be
simplified into one that does not rely on
approximations~\cite{brezak2011path}, except for evaluating Fresnel integrals.

Finally, in order to apply Theorem~\ref{theo-two-clothoids}, it
remains to choose values for $\kappa_1$ and $\kappa_2$ at the
extremities of interpolated curves. We use the following strategy:
\begin{itemize}
\item
At the junction between a straight line segment and an arc, or between
two arcs turning in opposite directions, a natural choice is $\kappa_i = 0$.
\item
If we need to connect two arcs turning in the same direction, with
respective curvatures $\kappa_{C1}$ and $\kappa_{C2}$ which we assume
w.l.o.g. to be positive, we have to choose a curvature $\kappa_i$ that
satisfies $\kappa_i < \min(\kappa_{C1}, \kappa_{C2})$. This can be
achieved by defining $\kappa_i = f \min(\kappa_{C1}, \kappa_{C2})$,
where $0 < f < 1$ is a reduction factor that can be arbitrarily
chosen. For the Eurobot application, we have observed that selecting $f
= 0.70$ leads to paths along which both the curvature and
variation of curvature stay within acceptable bounds.
\end{itemize}

\section{Speed Profile Computation}

Before addressing the computation of a speed profile for a given path,
we need to define the formalism in which such paths are
represented. As explained in
Section~\ref{sec-speed-profile-statement}, a discretized path takes
the form of a sequence $(x_0, y_0, \theta_0)$, $(x_1, y_1, \theta_1)$,
\ldots, $(x_m, y_m, \theta_m)$ of successive configurations of the
robot sampled at indices ranging from $0$ to $m$. The discretization
step between these configurations is usually much finer than for
specifying the broken lines that are input to the interpolation
procedure described in Section~\ref{sec-interpolation}. (In the
Eurobot application, the clothoid arcs synthesized by this procedure
are each typically discretized into dozens of intermediate
configurations.) This strategy differs from methods such
as~\cite{kunz12} and~\cite{scheuer04}, in which curves are
represented in analytic form. The advantages of our approach are that
discretized trajectories can be handled with much simpler data
structures, are not limited to curves that admit an analytic form,
and are less subject to numerical issues. We stress the fact that the
speed profile computation algorithm discussed in this section is not
restricted to the paths considered in Section~\ref{sec-interpolation},
but is applicable to arbitrary curves, in particular to those produced
by techniques such as~\cite{scheuer04}.

Let us now discuss the precise semantics of discretized paths.
Between a configuration $(x_i, y_i, \theta_i)$ and its successor
$(x_{i+1}, y_{i+1}, \theta_{i+1})$, we consider that the robot moves
along a circle arc, increasing its orientation by the angle
\mbox{$\delta_i = \theta_{i+1}-\theta_i$}, which is usually small in the
case of fine discretization. This circle arc
is fully characterized by the angle $\delta_i$ together with the chord
length
\[
\lambda_i = \sqrt{(x_{i+1} - x_i)^2 + (y_{i+1} - y_i)^2}.
\]
The
curvature $\kappa_i$ at this step thus satisfies
\[
\kappa_i =
\frac{2}{\lambda_i} \sin \frac{\delta_i}{2}.
\]
For the sake of simplicity, and in
order to be able to easily chain paths together, we impose this
curvature to be zero at the origin and endpoint of all paths.

Recall that the aim is to obtain a speed profile for a given path that
minimizes the total time needed for following this path.  We build
such a profile by computing the largest possible value for the speed
of the robot at each index from $0$ to $m$. This speed can potentially
be measured at various locations on the robot, such as its center of
mass, its wheels, \ldots\ For differential-drive, tricycle or car-like robots,
all those speeds can be expressed as functions of a single parameter
and the geometry of the path. A natural choice would be to define this
parameter as the speed of the robot measured at its center which, in
the case of differential-drive robot, is defined as the midpoint of
the line segment linking the two locomotion wheels. This solution
turns out to be problematic for reasoning about parts of paths where
the absolute curvature is high, which intuitively corresponds to
rotations of the robot around its center. In such a case, even though
the robot is in motion, its center moves only slowly, or not at all.

We choose instead to express all speeds of interest in terms of a
parameter $z_i$ that we call the \emph{velocity} at the current index
$i$ of the path, defined as the quadratic mean of the respective
speeds $v_{Li}$ and $v_{Ri}$ of the left and right wheels of the
differential drive:
\[
z_i = \sqrt{\frac{v_{Li}^2 + v_{Ri}^2}{2}}.
\]
This parameter has the advantage of being always positive, and nonzero
whenever the robot is not stationary. Knowing the geometry of the
robot, one can easily compute the speeds at its center, center of mass,
individual wheels, or other locations of interest,
from the velocity $z_i$ and the curvature $\kappa_i$ at the current
path index. When the curvature is zero, such as at the extremities
of paths, all those speeds become equal to $z_i$.

This reduces the speed profile computation problem to determining,
for each index $i$, the highest possible velocity $z_i$ at that point.
Note that one cannot realistically assume that this velocity remains constant
when the robot follows the circle arc from $(x_i, y_i, \theta_i)$ to
$(x_{i+1}, y_{i+1}, \theta_{i+1})$. Indeed, the speed of the robot would then
be discontinuous at the junction points between adjacent arcs, which would
complicate the handling of acceleration constraints.

A better solution is to consider that, between the indices $i$ and
$i+1$, the velocity varies linearly from $z_i$ to $z_{i+1}$ with
respect to traveled distance. With this assumption, all the
accelerations of interest (such as those measured at individual
wheels, at the center of mass or other locations, in the tangential or
radial directions, \ldots) can be derived from the values of $z_i$,
$z_{i+1}$, and $\kappa_i$, and take at all locations values that accurately
approximate those of the underlying (undiscretized) curve, assuming a
sufficiently fine discretization.

\begin{figure}
\vspace{5pt}
\centerline{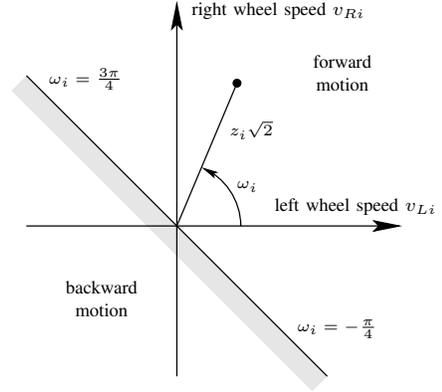}
\caption{Relations between speeds}
\label{fig-speeds}
\vspace{-5pt}
\end{figure}

The physical constraints imposed on the robot can be classified in two
groups. We first have constraints that translate into an upper bound
on the velocity $z_i$ at a path index $i$.  Let us illustrate this
situation with a simple example. Assume that the speeds $v_{Ri}$ and
$v_{Li}$ measured at respectively the right and the left wheels of the
differential drive (in the forward direction) are constrained to
belong to the interval $[-v_{i\mathit{max}}, v_{i\mathit{max}}]$ at
the path index $i$, with $v_{i\mathit{max}} > 0$.  Expressing the
wheel speeds in terms of $z_i$, we obtain
$v_{Ri} = z_i \sqrt{2}\,\sin \omega_i$ and 
$v_{Li} = z_i \sqrt{2}\,\cos \omega_i$,
where $\omega_i$ is defined as the angle that satisfies
\[
\displaystyle
\tan
\omega_i = \frac{1+\frac{e\kappa_i}{2}}{1 -\frac{e\kappa_i}{2}},
~~-\frac{\pi}{4} < \omega_i < \frac{3\pi}{4},
\]
and $e$ is the distance between the wheels. The relations
between $v_{Ri}$, $v_{Li}$, $z_i$ and $\omega_i$ are illustrated in
Figure~\ref{fig-speeds}.  Our constraint then gives out the upper
bound
\[
\begin{array}{c@{~~}l}
\displaystyle
z_i \leq \frac{v_{i\mathit{max}}}{\sqrt{2}\,\sin \omega_i}&
\displaystyle
\mbox{if
$\frac{\pi}{4} < \omega_i < \frac{3\pi}{4}$},\\[2ex]
\displaystyle
z_i \leq \frac{v_{i\mathit{max}}}{\sqrt{2}\,\cos \omega_i}&
\displaystyle
\mbox{if
$-\frac{\pi}{4} < \omega_i < \frac{\pi}{4}$}.
\end{array}
\]

The second group of constraints contains those that involve the
velocities $z_i$ and $z_{i+1}$ at two successive indices $i$ and
$i+1$. This group notably includes constraints expressed in terms of
accelerations. Let us give an example. The tangential acceleration
experienced at the center of a differential-drive robot
during step $i$ is given by
\[
a_{Ti} = \frac{v^2_{i+1} - v^2_i}{2s_i},
\]
where $v_i$ denotes the speed measured at the center at
the path index $i$, and $s_i = \delta_i/\kappa_i$ is the distance driven during
step $i$. Since we have 
$v_i = (v_{Ri} + v_{Li})/2$
for all $i$, this expression becomes
\[
a_{Ti} =
\frac{(\sin \omega_{i+1} + \cos \omega_{i+1})^2 z_{i+1}^2 -
(\sin \omega_i + \cos \omega_i)^2 z_i^2}{4s_i},
\]
where $\omega_i$ is defined as in the previous example. One easily
sees that imposing bounds on this acceleration amounts to enforcing a
constraint over both $z_i$ and $z_{i+1}$.  At all every path index $i
\in [0, m-1]$, we denote by $\phi_i(z_i, z_{i+1})$ the conjunction of
all physical constraints that jointly involve $z_i$ and
$z_{i+1}$. Note that constraints in both groups may be
context-sensitive: The constraints involving $z_i$ and $z_j$ at two
different locations $i$ and $j$ can differ, or be expressed with
respect to different values of parameters. This makes it possible, for
instance, to impose tighter speed limits in the close vicinity of
obstacles.

We are now ready to describe our procedure for computing the fastest
physically-feasible speed profile for a given path.  This procedure
has originally been introduced in~\cite{lens2008tfe}. It differs
from~\cite{brezak2011time,kunz12} by the much broader range of
physical constraints that it supports.  The algorithm
proceeds in three stages. In the first one, it computes for every
index $i$, the largest possible value $z_{i\mathit{max}}$ allowed by
the constraints of the first group at that point.  It is also
essential to make sure that the constraints that belong to the second
group remain satisfiable: if $z_{i\mathit{max}}$ is such that the
constraint $\phi_i(z_{i\mathit{max}}, z_{i+1})$ does not hold for 
at least one $z_{i+1} \leq z_{i+1\mathit{max}}$, then $z_{i\mathit{max}}$ 
has to be lowered into the largest value that makes the constraint 
satisfiable. In the same way, $z_{i\mathit{max}}$ must be sufficiently 
small for the constraint $\phi_{i-1} (z_{i-1}, z_{i\mathit{max}})$ 
to be satisfiable in $z_{i-1}$. These operations can be carried out 
numerically, a simple strategy being to perform a binary search until 
the required precision is reached.

After a suitable value of $z_{i\mathit{max}}$ has been obtained at all
indices $i$, the second stage assigns a tentative value to
$z_i$ for increasing values of $i$. The initial velocity $z_0$ is
fixed by the speed of the robot specified at the origin of the
path. Then for $i = 1, 2, \ldots$, we successively compute the largest
value of $z_i$ that is less than or equal to the upper bound
$z_{i\mathit{max}}$ at the index $i$, and that satisfies the
constraint $\phi_{i-1}(z_{i-1}, z_i)$ (with the value of $z_{i-1}$
obtained at the previous step).

The third and last stage performs a similar operation for decreasing
values of $i$, starting from the last index $m$ of the path and moving
towards its origin. Before the first iteration, the velocity $z_m$ at
the end of the path has possibly to be lowered in order to satisfy the
upper bound imposed on the final speed of the robot. Then, for $i =
m-1, m-2, \ldots$, one successively adjusts the current value of $z_i$
(by lowering it or leaving it unchanged) so as to satisfy the
constraint $\phi_{i}(z_{i}, z_{i+1})$.

After completing this stage, the computed values of $z_i$ at all path
indices are such that constraints of both groups are satisfied, and it
remains to check that the computed initial velocity $z_0$ corresponds
to the initial speed of the robot. In the case of a mismatch (meaning
that $z_0$ had to be lowered during third stage), it is impossible to
follow the path with the specified initial speed while satisfying all
physical constraints, and the speed profile computation returns an
error. Otherwise, one can straightforwardly compute, from the value of
$z_i$ at all indices and the geometry of the path, the instant $t_i$ at
which the corresponding configurations will be reached.

This technique yields a time-optimal speed profile, since increasing
the speed of the robot at any location on the path would lead to
violating at least one physical constraint. The computational cost is
linear in the number of path steps, provided that all constraints can
be solved in bounded time.

\section{Conclusions}

In this paper, we have addressed the problem of interpolating a path
expressed as a sequence of straight line segments into a trajectory
that can be followed as fast as possible by a nonholonomic robot, taking
into account the physical constraints of the robot.  This problem has
been well studied in the
literature~\cite{latombe1991planning,choset2005principles,lavalle2006planning,
scheuer04,kuwata09,kunz12}, but our motivation for
developing an original solution was prompted by the particular
requirements of the Eurobot contest. In this setting, it is crucial to
be able to plan trajectories that can be generated with very low
computational cost, to take into account complex physical constraints
such as those governing traction at individual wheels or ensuring
stability in turns, and to provide accurate spatial and temporal
advance information about the visited configurations. To the best of our
knowledge, our solution is the first one that meets all those
requirements. Compared with methods such
as~\cite{shin1990path,scheuer04} that also
rely on clothoids for interpolating paths, our approach of joining
only two arcs of clothoids for moving from one curvature to another
has the advantage of being simpler and computationally cheaper, the
trade-off being that the generated curves are not guaranteed to be
optimal.

The path interpolation and speed profile computation algorithms
discussed in this paper have been implemented in the robots built for
Eurobot at the University of Liège since 2008, together with an
original path planning algorithm. In this setting, they have been
successfully validated on hundreds of thousands of trajectories,
considering 16 distinct physical constraints of robots with
differential as well as tricycle drive, some of them being
context-sensitive: lower and upper bounds on the speed and
acceleration of locomotion and steering wheels, on the speed, angular
speed, tangential and radial accelerations at the center of mass,
and on the angular rate of steering. In order to illustrate the
efficiency of our method, we report in Figure~\ref{fig-runtimes} the
time needed for running the interpolation and speed profile algorithms
on a few sample trajectories experienced in the Eurobot
application. We distinguish the costs of the curve synthesis, path
discretization, and speed profile computation steps. The total
computational cost typically amounts to less than half a
millisecond of CPU time on a i5-460M processor running at 2.53 GHz,
which is several orders of magnitude faster than techniques such
as~\cite{kunz12}.

\begin{figure}
\footnotesize
\vspace{5pt}
\begin{tabular}{lllll}
\\\hline
Discretized & Synthesis & Discretization & Speed profile & Total \\
points    & time      & time           & time          & time  \\
\hline
292 & 15.2 $\mu$s & 33.4 $\mu$s & 140.9 $\mu$s & 189.5 $\mu$s\\
520 & 29.6 $\mu$s & 53.6 $\mu$s & 239.0 $\mu$s & 322.2 $\mu$s\\
632 & 30.8 $\mu$s & 72.7 $\mu$s & 303.5 $\mu$s & 407.0 $\mu$s\\
656 & 23.1 $\mu$s & 77.9 $\mu$s & 319.1 $\mu$s & 420.1 $\mu$s
\\\hline
\end{tabular}
\caption{Experimental results}
\label{fig-runtimes}
\vspace{-10pt}
\end{figure}

\bibliographystyle{IEEEtranS}
\bibliography{biblio}

\end{document}